\documentclass{ecai}
\usepackage{graphicx}
\usepackage{latexsym}

\usepackage{times}
\usepackage{soul}
\usepackage{url}
\usepackage[hidelinks]{hyperref}
\usepackage[utf8]{inputenc}
\usepackage[small]{caption}
\usepackage{graphicx}
\usepackage{amsmath}
\usepackage{amsthm}
\usepackage{booktabs}
\usepackage{algorithm}
\usepackage{algorithmic}
\usepackage[switch]{lineno}

\usepackage{textcomp}
\usepackage{xcolor}
\usepackage{subfigure}
\usepackage{diagbox}
\usepackage{threeparttable}
\usepackage{multirow}
\usepackage{color}
\usepackage{bbold}
\usepackage{setspace}
\usepackage{fontawesome}

\begin{document}

\begin{frontmatter}

\title{Individual Fairness under Uncertainty}



\author[A]{\fnms{Wenbin}~\snm{Zhang}\thanks{\faEnvelope~ wenbin.zhang@fiu.edu}}
\author[A]{\fnms{Zichong}~\snm{Wang}}
\author[B]{\fnms{Juyong}~\snm{Kim}\textsuperscript{\normalfont b}, \\\fnms{Cheng}~\snm{Cheng}}
\author[C]{\fnms{Thomas}~\snm{Oommen}} 
\author[B]{\fnms{Pradeep}~\snm{Ravikumar}}
\author[D]{\fnms{Jeremy}~\snm{Weiss}}

\address[A]{Florida International University, Miami, FL 33199}
\address[B] {Carnegie Mellon University, Pittsburgh, PA 15213}
\address[C]{University of Mississippi, Oxford, MS, 38677}
\address[D]{National Institutes of Health, Bethesda, MD 20892}

\begin{abstract}
Algorithmic fairness, the research field of making machine learning (ML) algorithms fair, is an established area in ML. As ML technologies expand their application domains, including ones with high societal impact, it becomes essential to take fairness into consideration during the building of ML systems. Yet, despite its wide range of socially sensitive applications, most work treats the issue of algorithmic bias as an intrinsic property of supervised learning, \emph{i.e.}, the class label is given as a precondition. Unlike prior studies in fairness, we propose an individual fairness measure and a corresponding algorithm that deal with the challenges of uncertainty arising from censorship in class labels, while enforcing similar individuals to be treated similarly from a ranking perspective, free of the Lipschitz condition in the conventional individual fairness definition. We argue that this perspective represents a more realistic model of fairness research for real-world application deployment and show how learning with such a relaxed precondition draws new insights that better explains algorithmic fairness. We conducted experiments on four real-world datasets to evaluate our proposed method compared to other fairness models, demonstrating its superiority in minimizing discrimination while maintaining predictive performance with uncertainty present.
\end{abstract}

\end{frontmatter}

\section{Introduction}

There is recent concern that we are in the midst of a discrimination crisis within the field of machine learning (ML) and artificial intelligence (AI)~\cite{saxena2023missed,zhang2021fair,zhang2023fairness,zhang2022fairness,zhang2023censored}. Rightfully, the AI/ML community has conducted vast research to study the quantification and mitigation of algorithmic bias, which is critical for the use of algorithmic decision-making systems in domains of high societal impact such as criminal justice~\cite{chouldechova2017fair}, healthcare~\cite{chen2020ethical}, predictive policing~\cite{wang2023fg2an}, and employment~\cite{wang2023towards}. Thus far, most studies 
tackle the problem by proposing fairness constraints via regularizers/optimizations at the group level: first identify a \textit{sensitive attribute}, \textit{e.g.}, race or gender, as a potential source of bias among the collection of high-level groups; then achieve parity for some fairness statistics of the classifier, \textit{e.g.} the prediction accuracy and true positive rate, across the predefined groups~\cite{mehrabi2021survey}. These group fairness approaches, however, are inapplicable when class label uncertainty is present~\cite{zhang2022longitudinal}. Additionally, while group fairness enjoys the merit of easy operationalization, its aggregative characteristic makes it easy to fail~\cite{barocas2017fairness}.

In contrast, the \textit{individual fairness} approach alleviates these drawbacks by evaluating a finer granularity of fairness at individual level. 
The compelling notion of individual fairness is proposed in the seminal work of~\cite{dwork2012fairness}, which requires similarly situated individuals to receive similar probability distributions over class labels to prevent inequitable treatment. Individual fairness, without the need to explicitly identify sensitive attributes, is much less restrictive than group fairness. 
However, the Lipschitz condition required in existing individual fairness literature is a nontrivial task to satisfy, 
as 1. the Lipschitz constant specification is hard due to the difference in distance metrics between the input and outcome spaces;
2. distance calibration is required as the absolute distance comparison in the Lipschitz condition tends to fail in calibrating differences among different individuals~\cite{dong2021individual}.
Such difficulty was also pointed out in \cite{mukherjee2020two} but only resulted in additional efforts of metric learning, whereas our rank-based method removes the need of the Lipschitz constant and distance calibration by avoiding the absolute distance value comparison.

\begin{figure}[t]
	\centering
	\includegraphics[width=0.46\textwidth]{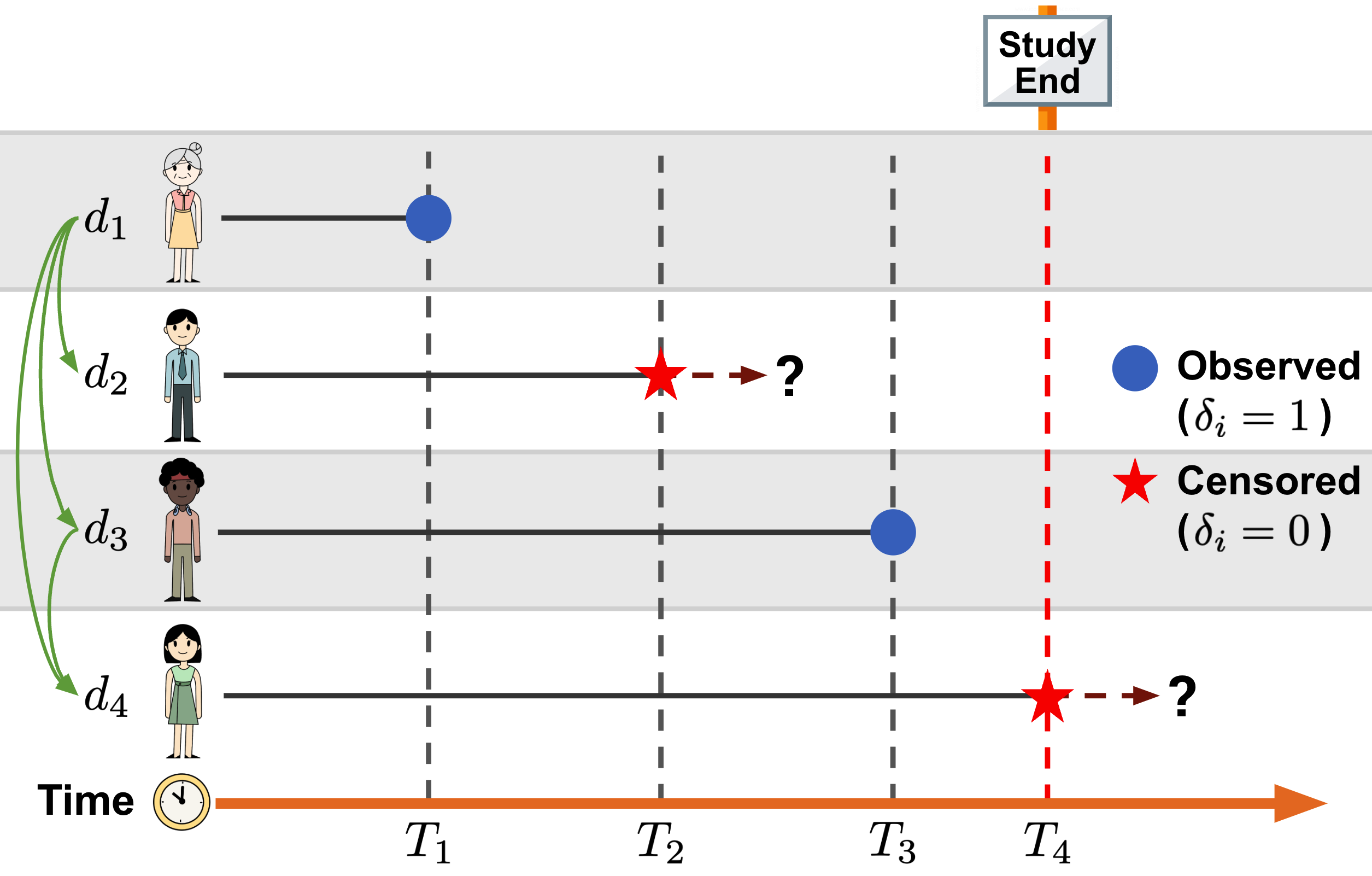}
	\vspace{-0.2cm}
	\caption{An illustration of the censoring phenomenon. Individuals $d_2$ and $d_4$ are censored while others, \textit{i.e.}, $d_1$ and $d_3$, are non-censored. Individuals are arranged in the increasing time order of their survival times with the lowest, $i.e$, $T_1$, being at topmost. The study ends at the time shown as the red vertical dash line. There is no edge originating from a censored individual due to censorship, which means that pair comparison between two individuals cannot be made when the individual with lower survival time is censored.}
	\label{fig:toyexample}
\end{figure}

Another major obstacle in the real-world applicability of individual (and also group) fairness is the assumption of full class label availability, which fails when there is uncertainty in class labels due to \textit{censoring}, a phenomenon where the information about the event of interest is partially known~\cite{zhang2022longitudinal,kvamme2019time,turner2022longitudinal,liu2021research,guyet2022incremental}. 
Considering an example from a clinical prediction task (Figure~\ref{fig:toyexample}), for censored individuals $d_2$ and $d_4$, the true time to relapse or hospital discharge is unknown, causing the uncertainty in class labels.
Due to the inability to handle censorship information, existing fairness studies quantify and mitigate bias by focusing on the proportion of data with assured class label, thus either dropping observations with uncertain class labels~\cite{chouldechova2017fair,du2021learning,zhao2019rank} or omitting the censorship information~\cite{wan2020denoising,vasudevan2020lift,zhang2019faht}. However, removing them would bias the results towards the individuals with known class labels~\cite{turkson2021handling,zhang2021disentangled,zhang2018deterministic}.

In summary, there is a need for an algorithm that addresses individual fairness in ML under uncertainty, an under-explored area of research, with two requirements: \textbf{i) Free from the Lipschitz condition resulting from the principle of individual fairness.} Without this, the algorithm may have limited use cases due to the metric calibration between the input and output spaces. \textbf{ii) Quantifying and mitigating bias under uncertainty.} The algorithm should not ignore the uncertainty in censored data or the censorship information to avoid bias.

To tackle the aforementioned issues, this paper conducts an initial investigation of \textit{individual fairness under uncertainty} for a fairness guarantee more in line with realistic assumptions across individuals and free from the Lipschitz condition. Our individual fairness measure, named \textit{Fair Normalized Discounted Cumulative Gain} (\textit{FNDCG}), is motivated by the same individual fairness principle~\cite{dwork2012fairness} that similar individuals should be treated similarly, while formulated as the correlation of similarities in the feature and risk spaces respectively, establishing a new fairness measure usable on censored data. Along with FNDCG, we also propose a corresponding algorithm to address discrimination involving censored individuals. Our method, named \textit{fairIndvCox}, augments the standard model of survival analysis, the Cox proportional hazard model, by being aware of individual fairness while learning the parameters of risk prediction.

To our knowledge, this work is the first attempt to quantify and mitigate bias under the individual fairness principle, but from a ranking perspective, with uncertainty present, and \textit{free of} the Lipschitz condition. Our major contributions are summarized as follows:

\begin{itemize}
    \item We formulate a new research problem of individual fairness guarantee in learning with uncertainty. 
    \item We devise \textit{FNDCG}, a new notion of individual fairness to measure bias on censored data. Defined with the correlation of similarity in the feature space and the one in the risk prediction space, FNDCG does not require Lipschitz condition and complete class labels.
    \item We propose a debiasing algorithm named \textit{fairIndvCox} for bias mitigation in censorship settings, by incorporating our individual fairness measure into the standard model of survival analysis.
    \item We evaluate our debiasing algorithm on four real-world datasets with censorship, comparing it with four survival analysis algorithms and its Lipschitz variant. This confirms the utility of the proposed approach in practice. Further analysis also illustrates the trade-off between individual fairness and predictive performance.
\end{itemize}

The remainder of this paper is organized as follows. In Section~\ref{sec:related_work}, we describe related work in fair machine learning and learning with uncertainty, followed by the preliminaries of survival analysis and the problem definition in Section~\ref{sec:notations}. In Section~\ref{sec:method}, we propose our notion of individual fairness under uncertainty and corresponding survival model with an individual fairness specification. In Section~\ref{sec:experiments}, we empirically validate the effectiveness of our learning algorithm on real-world survival analysis datasets and provide qualitative analysis on the effect of the hyper-parameters on the model. Finally, we conclude and provide future directions in Section~\ref{sec:conclusion}.

\section{Related Work}
\label{sec:related_work}


\subsection{Censored Data}

In many real-world applications, the main outcome under assessment, \textit{i.e.}, the class label, could be unknown for a portion of the study group. 
This phenomenon, deemed censorship, can arise in various ways, hindering the use of many algorithms. For example~(Figure~\ref{fig:toyexample}), a study may end before an individual experiences the event of interest, \textit{e.g.}, individual $d_4$. The studied individual can also be lost to follow-up during the study period, withdraw from the study, or experience a competing event making further follow-up impossible, \textit{e.g.}, individual $d_2$. In the typical setting of survival analysis, censored examples are only guaranteed not to have experienced events until their last observation, \textit{e.g.} $t_2$ and the end of the study for $d_2$ and $d_4$, respectively, and we do not know their exact class labels.

The censorship information is used together with the observed data to fit or evaluate survival models, a statistical model that analyzes the expected duration of time until each individual's event. Specifically, we can guarantee that a censored example with the time of event $T$ happens after $T$, so we can compare two events at $T_1$ and $T_2 $ for $T_1 < T_2$ if neither is censored at $T_1$, regardless of censorship at $T_2$. For instance, the green edges in Figure~\ref{fig:toyexample} represent the comparable pairs among individuals with censored and observed events (as the order graph), from which we can tell that $d_1$ happens before $d_2$, while whether $d_2$ happens before $d_3$ or not remains unknown.

Given that censored data is common, \textit{e.g.}, clinical prediction (Support)~\cite{knaus1995support}, marketing analytics (KKBox)~\cite{kvamme2019time}, recidivism prediction instrument datasets (COMPAS~\cite{angwin2016there} and ROSSI~\cite{fox2012rcmdrplugin}), survival analysis has gained popularity in applied work. For example, in customer analytics whether a customer will cancel the service, \textit{e.g.,} event of interest/class label, can be unknown due to various reasons discussed above~\cite{kvamme2019time}. Similarly, one may predict in domains of reoffense~\cite{angwin2016there}, analyzing financial outcomes in actuarial analysis~\cite{wang2019machine}, and predictive maintenance in mechanical operations~\cite{wang2021harmonic}. 


\subsection{Fairness in AI}

\paragraph{Quantifying Bias}
Much progress has been made to quantify and mitigate unfair or discriminatory behaviours of AI algorithms. These efforts, at the highest level, can be typically divided into two families: \textit{individual fairness} and \textit{group fairness}. A vast majority of existing works focuses on the group notions, aiming to ensure members of different groups, \textit{e.g.,} gender or race \textit{aka} sensitive attributes, achieve approximate parity of some statistics over class labels, such as statistical parity~\cite{zhang2019faht}, disparate impact~\cite{wang2023fg2an}, and equality of opportunity~\cite{wang2023mitigating}. While enjoying the merit of easy operationalization, group-based fairness methods fail at guaranteeing fairness at the individual level in addition to several other drawbacks~\cite{barocas2017fairness}. 

Individual fairness, on the other hand, alleviates such a drawback by requiring that individuals who are similarly situated with respect to the task at hand receive similar probability distributions over class labels~\cite{dwork2012fairness}. Formally, this objective can be formulated as the Lipschitz property, and fairness is thus achieved iff:

\begin{equation}
	\label{equ:lipschitz}
	D(f(x_a), f(x_b)) \leq LD'(x_a, x_b)
\end{equation}

\noindent where $L$ is the Lipschitz constant, $D'(\cdot, \cdot)$ and $D(\cdot, \cdot)$ are corresponding distance functions of features in input space, $x$, and probability distributions over class labels in output space, $f(\cdot)$, respectively.
The major obstacles for wider adoption of individual fairness, though, are the difficulty of calibrating the distance functions resulted from the Lipschitz condition and the assumption of the availability of class labels, which is impractical in many applications due to censorship. For instance, in the ML-task of predicting critical illness in COVID-19 patients \cite{liang2020development},
clinical knowledge is required to calibrate the distance-based comparison in Equation~\ref{equ:lipschitz} since a 10-year difference in age ($D'(\cdot, \cdot)$) for patients younger than 25 would likely result in not much of a difference in risk outcomes ($D(\cdot, \cdot)$), whereas a 10-year difference ($D'(\cdot, \cdot)$) for patients older than 65 could lead to a significant increase in the risk of progressing to critically ill ($D(\cdot, \cdot)$). In addition, a patient may experience censorship, introducing uncertainty about the true progression of their illness at the time of evaluation.

Our new individual fairness methodology resolves these two main limitations in current literature, providing a fairness guarantee across individuals with censorship and is free from the Lipschitz condition. 

\paragraph{Mitigating Bias}
The fairness notions mentioned above are used as a constraint or as a regularizer to enforce fairness. These debiasing algorithms, mostly group-based, can be categorized into three groups based on the stage where machine learning intervention happens: the pre-processing, in-processing, and post-processing groups.

The first group, pre-processing approaches, works on bias in the data or input stage, assuming that unbiased training data is accessible for a fair ML model. These methods modify the data distribution to ensure fairness of the representations from different groups and are model-agnostic. Examples of this group include data massaging~\cite{kamiran2009classifying}, which changes data distribution, an extension called local massaging~\cite{vzliobaite2011handling}, and reweighing~\cite{calders2009building}, which assigns different weights to the communities.

The second group, in-processing approaches, directly changes ML algorithms to produce unbiased predictions and is generally model-specific. For example, in~\cite{zhang2019faht}, the fairness gain is incorporated into the splitting criteria of the Hoeffding Tree algorithm, which is later extended in~\cite{zhang2021farf} to ensemble-based methods. In~\cite{zhao2019rank}, the Mann Whitney U test is applied to fairness learning in multi-task regression. These methods focus on group fairness and require complete class labels. Yet, there is a limited number of research on individual fairness under data censorship, which this work focuses on.

The last group, post-processing approaches, modifies the decision boundaries to fairly represent diverse groups. 
Examples include building an interpretable model~\cite{zeng2017interpretable}, adjusting the decision threshold to reduce unfairness~\cite{hardt2016equality}, and moving decision boundaries of the deprived communities to prevent discrimination~\cite{fish2016confidence}. However, applying these techniques under censorship is not straightforward, as decision boundaries may also be censored owing to their distribution.

\subsection{Survival Analysis}
\label{sec:survivalAnalysis}

The prevalence of censored data motivates the study of survival analysis to address the problem of partial survival information from the study cohort~\cite{clark2003survival}. The Cox proportional hazard (CPH) model \cite{cox1972regression} is the most commonly used method, which expresses the hazard function as the product of a shared time-dependent baseline hazard and an individual-specific risk function. Developing the CPH model, \cite{katzman2018deepsurv} parameterized the effect of an individual's covariates by a neural network. Another line of research is tree-based methodology~\cite{bou2011review,ishwaran2008random}, where the splitting rule is modified to handle censored data and is free from the proportional assumption of the CPH model. Interested readers may refer to \cite{wang2019machine} for a comprehensive survey on recent methods of modeling censored data.

Like other AI approaches, care must be taken to ensure the fairness of survival models to prevent bias against deprived communities. Starting with~\cite{zhang2022longitudinal}, there is a line of work studying fairness with censorship but subject to group-based constraints. In addition, the survival model is modified to ensure fair risk predictions as in~\cite{keya2021equitable}. However, their work requires the Lipschitz condition as in conventional individual fairness and does not explicitly consider the survival information to address discrimination. Our method aims to address these two limitations.

\section{Notations and Problem Definition}
\label{sec:notations}

In this section, we provide preliminary notations and concepts of survival analysis, followed by the definition of the problem of our concern. In survival analysis, censored data can be typically described as follows. Each individual $d_i$ with index $i \in \{1, \cdots, N\}$ is equipped with a characteristics tuple $(x_i, T_i, \delta_i)$, where the entries of each tuple are i) $x$: the observed feature, ii) $T$: the survival time, \textit{i.e.} the time of event, and iii) $\delta$: the event indicator, which indicates whether the event is observed. In the setting of survival analysis, the event is observed only when $\delta=1$, and $T$ is the actual time of event. When $\delta = 0$, the event time is censored, resulting in uncertainty on the class label and is only known to be greater than or equal to $T$.

The modeling function commonly used is the \textit{hazard function}, which specifies the instantaneous rate of event occurrence at a specified time $t$ conditioned on surviving to $t$: 
\begin{equation}
	h(t|x) = \lim\limits_{\bigtriangleup t \rightarrow 0} \frac{\text{Pr}(t<T<t+\bigtriangleup t| T\geq t, x)}{\bigtriangleup t}
\end{equation}
\noindent Given a hazard model, one can also compute the survival function $S(t|x) = \text{Pr}(T>t|x)$, the probability that the event occurs after a specific time $t$ by
 \begin{equation}
	S(t|x) = \exp\left(-\int_0^t h(t|x) dt\right)
\end{equation}
Among the various proposed survival analysis methods, the Cox proportional hazards model (CPH)~\cite{cox1972regression} has become the standard for modeling censored data, which defines the relation between the hazard function $h(t|x)$ and the covariates as:
\begin{equation}
	h(t|x) = h_0(t)\exp(\beta^\top x)
\end{equation}
where $h_0(t)$ is called the baseline hazard function (\textit{i.e.}, when $x=0$), and $\beta$ is a set of unknown parameters which can be estimated by applying the maximum likelihood estimation. Given a dataset of $N$ individuals $\{(x_i, T_i, \delta_i)\}_{i=1}^N$ with i.i.d. assumption, we can compute the likelihood as the product of the likelihood of the uncensored individuals. Such function is called the partial likelihood and can be written as follows:
\begin{equation}
	\label{equ:likelihood}
	L(\beta)= \prod_{i:\delta_i=1}  \frac{\exp(\beta^\top x_i)}{\sum_{j:T_j\geq T_i}\exp(\beta^\top x_j)}
\end{equation}
\noindent The partial likelihood estimate $\hat{\beta} = \arg\max_\beta L(\beta)$ can be obtained by maximizing the partial likelihood function. Note the partial likelihood function does not include the baseline hazard function. One can also add a regularization function, such as ridge or lasso regularization, for $\beta$.

To evaluate survival models, the \textit{concordance index}, or \textit{C-index}, is commonly used~\cite{harrell1982evaluating}. Given a survival model, the C-index of the model measures the fraction of all comparable pairs of individuals whose predicted survival times are correctly ordered as training data:
\begin{equation}
    C = \frac{1}{\!\sum\limits_{i:\delta_i=1} \!\!\! |\{j:T_j > T_i \}|} \!\! \sum\limits_{i:\delta_i=1} \sum\limits_{j:T_j > T_i}  \!\!\! \mathbb{1} [f(x_j) > f(x_i)]
\end{equation}
\noindent where $f(x)$ is the expected survival time for an individual~\cite{steck2007ranking}. C-index is also equal to the area under the ROC curve (AUC) in the presence of censorship. In a proportional hazard model, the order of expected survival time is the same as the order of the hazard function. Please see Figure 1 for an example of an order graph, which represents the comparable pairs of individuals. 

The main problem we address in this work is to devise an algorithm that can quantify the individual fairness notion in survival analysis and use the quantification to mitigate the bias. Under the general assumption of survival analysis, unlike most existing works of individual fairness, not all individuals are given a label, or survival time, due to data censoring. Another desirable quality the algorithm has is to alleviate or be free from the Lipschitz condition, enabled by the locality between similarity metrics. Note that although similarity-based constraints have been formulated to alleviate bias~\cite{kang2020inform,dong2021individual}, we are the first to make the contribution of taking censored information into consideration while establishing our similarity-based constraint.

\section{Method}
\label{sec:method}

We introduce a learning algorithm for individual fairness with censored data. In Section~\ref{sec:notion}, we define a rank-based similarity measure of risk scores and propose a corresponding individual fairness score, named \textit{FNDCG@k}. In Section~\ref{sec:algorithm}, we propose a survival analysis model, named \textit{fairIndvCox}, which incorporates FNDCG@k into the Cox proportional hazard model.

\subsection{Individual Fairness Notion under Uncertainty}
\label{sec:notion}

Existing individual fairness notions depend on the Lipschitz condition, which is non-trivial due to the difference in the similarity metrics of the input and output spaces. In addition, they do not consider survival information when quantifying unfairness, which is important and requires special attention; otherwise, substantial bias could be introduced. To overcome these, we propose to evaluate unfairness from a ranking perspective while jointly considering survival information.

For each individual, we obtain two ranked lists of other individuals based on the similarity matrix $\text{Sim}_{D'}$ (on the input space) and $\text{Sim}_D$ (on the output space), and require the relative \textit{order} of individuals in these two lists to be consistent with each other. 
The intuition still follows conventional individual fairness as similar individuals should have similar prediction results, but approaching it from a ranking perspective instead of the absolute distance value comparison (Equation~\ref{equ:lipschitz}) promotes applicability by avoiding Lipschitz specification and distance calibration.
For instance (Figure~\ref{fig:toyexample}), assume the ordered list derived from $\text{Sim}_{D'}$ between patient $d_1$ and three other patients is \{$d_3$, $d_2$, $d_4$\}, ordered by closest-to-farthest.
Then, the predictions are individually fair for $d_1$ if the encoded list from $\text{Sim}_D$ is \{$d_3$, $d_2$, $d_4$\} as well, \textit{i.e.} fairness is obtained when patients, ordered by their similarity to patient $d_1$, have predicted risks in the same order of similarity to $d_1$'s risk. This potentially results in a patient more similar to $d_1$ receiving a more similar treatment as $d_1$.   
Note that the input similarity matrix $\text{Sim}_{D'}$ is often given a priori as it is problem-specific~\cite{lahoti13operationalizing,lahoti2019ifair}, while we define $\text{Sim}_D$ as follows,

{  
\setlength{\abovedisplayskip}{0pt}
\begin{align}
    \label{equ:similarity}
    \text{Sim}_{D,ij} & = \text{Sim}_{D}(d_i, d_j) = \exp\left( -|\bar{h} (t|x_i)- \bar{h} (t|x_j)| \right) \nonumber \\
    & = \exp\left(-|\exp(\beta^\top  x_i) - \exp(\beta^\top  x_j)| \right)
\end{align}
}

\noindent $\text{Sim}_{D,ij}$ is the $(i,j)$-th entry of $\text{Sim}_D$ and $\bar{h}(t|x)$ is the hazard function with $h_0(t)$ dropped, \textit{i.e.}, $\bar{h}(t|x) = \exp(\beta^\top x)$, as it is not individually specific in the CPH model.

In Equation~\ref{equ:similarity}, the similarity metric is formulated as the exponential of the negative difference of the risk score. We make a note that this considers various factors to make a similarity metric that performs a trade-off between accuracy and fairness. First, the exponential followed by negation is used for smoothing. This bounds the difference in the unbounded risk scores to a value between 0 and 1. Second, it transforms a distance metric into a similarity function, which has a value closer to 1 when the two individuals are similar. It also makes the function applicable to discounted cumulative gain (DCG)~\cite{burges2006learning}, which will be used to compute the fairness quantification. In DCG@k, the quality of the most similar pairs in the output space will be accumulated with a discounted factor decaying with their ranking. Here, similarity is more proper than a metric for the quality function as the closer a pair is, the higher the function is.

Since the encoded ranking list should also take important survival information and consistency between predicted and actual outcome into consideration, we adjust $\text{Sim}_D$ according to the concordance difference ($C_\triangle$):
\begin{equation}
	\text{Sim}_{D,ij} = (1- C_\triangle(x_i, x_j))\exp\left(-|\exp(\beta^\top  x_i) - \exp(\beta^\top  x_j)|\right)
 \label{equ:sim}
\end{equation}

\noindent where $C_\triangle(x_i, x_j)=|C_{x_i}- C_{x_j}|$ measures the concordance difference between $x_i$ and $x_j$. The concordance of individual $x_g$ within the ranking list, $C_{x_g}$, is defined as: 
\begin{align}
    C_{x_g} & = \frac{1}{\sum_{g'\neq g}\mathbb{1}[\delta_{<}=1]}\sum_{g'\neq g} \mathbb{1}[h(t|x_>)< h(t|x_<), \delta_{<}=1] \nonumber \\
            & = \frac{1}{\sum_{g'\neq g}\mathbb{1}[\delta_{<}=1]} \nonumber \\
            & ~~~~~~~ \times \sum_{g'\neq g} \mathbb{1}[\exp(\beta^\top  x_{>}) < \exp(\beta^\top  x_{<}), \delta_{<}=1]
\end{align} 

\noindent where 
$x_>$ and $x_<$ are the individuals with a longer, \textit{i.e.} $T_>= \max(T_g, T_g')$, and a shorter, \textit{i.e.} $T_<= \min(T_g, T_g')$, survival time, and $\delta_{<}$ is the event indicator of shorter survival time. $C_{x_g}$ can be interpreted as the fraction of all other individuals whose predicted survival times are correctly ordered with $x_g$ considering their actual survival times. The concordance difference effectively adjusts the similarity values defined in Equation~(\ref{equ:similarity}) by penalizing the cases where one individual's predicted survival time aligns well with that of others, while another individual's does not. In the general case, we would like the original similarity in the output space to be downscaled according to the prediction deviation as reflected by the concordance difference, which also explicitly includes survival information when quantifying unfairness in the censoring setting.

Armed with the similarity matrix $\text{Sim}_D$ and $\text{Sim}_{D'}$, we propose the \textit{Fair Normalized Discounted Cumulative Gain (FNDCG@k)}, motivated by learning to rank~\cite{burges2006learning}, as a metric for the evaluation of individual fairness with censorship defined as follows:
\begin{equation}
	\label{equ:FNDCG@k}
	\text{FNDCG@k} = \frac{1}{N}\sum_{n=1}^{N}\frac{\text{DCG}_{\text{Sim}_D(d_n)}}{\text{DCG}_{\text{Sim}_{D'}(d_n)}}
\end{equation}

\noindent where $N$ is the number of individuals and $\text{DCG}_{\text{Sim}(d_n)}$ is the discounted cumulative gain of $d_n$ defined as: 


\begin{equation}
	\text{DCG}_{\text{Sim}(d_n)}= \sum_{\text{pos}=1}^{k} \frac{\text{Sim}_{D'}(d_{l_\text{pos}},d_n)}{\log(\text{pos}+1)}
\end{equation}

\noindent where $k$ is the length of the ordering list, $\{l_\text{pos}\}_{\text{pos}=1}^k$ is the ordering list of individual indices derived from the similarity matrix $\text{Sim}$ for individual $d_n$, and $\text{Sim}_{D'}(d_{l_\text{pos}},d_n)$ is the similarity in the input space between the individual at the $\text{pos}$-th position of the ordering list, $d_{l_\text{pos}}$, and the individual $d_n$. It is important to note that both $\text{DCG}_{\text{Sim}_D(d_n)}$ and $\text{DCG}_{\text{Sim}_D'(d_n)}$ are computing the DCG of the similarity values from $\text{Sim}_{D'}$, and the corresponding similarity is used only for deriving the ordering list $l_\text{pos}$. 

Essentially, FNDCG@k computes the per-individual average ratio between the DCG of input space similarity ranked by output space similarity and the maximum achievable DCG of input space similarity. Therefore, the value of FNDCG@k is within the interval of [0,1], which aligns with the existing individual fairness notions. In addition, the higher the FNDCG@k score, the more consistency between the ranking list encoded from the input and output space and thus, the fairer the model. 

The intuition behind enforcing FNDCG@k lies in promoting the consistency between the two ranking lists from the input and output spaces, i.e. having individuals ranked closer in the input space (\textit{e.g.} similar clinical condition) ranked closer in the output space (\textit{e.g.} similar risks and thus similar allocation of medical resources).
Moreover, focusing on the top-k ranking promotes local similarity without enforcing global similarity, corresponding to the individual fairness concept of promoting similar outcomes for similar individuals instead of for a group of individuals.
Finally, it is worth mentioning that our ranking perspective of individual fairness also possesses the potential of generalization to applications with full label availability, leaving the possibility of future expansion.

\subsection{Individual Fairness Algorithm under Uncertainty}
\label{sec:algorithm}

With the tailored individual fairness definition 
accounting for censoring, we now introduce a corresponding learning algorithm, \textit{fairIndvCox}, following the classic Cox proportional hazard model for modeling censored data, to generate tailored forecasts while providing fair risk predictions across individuals. Essentially, the learning algorithm augments the partial likelihood maximization of the CPH model with our individual fairness quantification, FNDCG@k.

Starting with the model utility maximization, the utility loss function $\mathcal{L}_\text{utility}$ is formulated as the negative log partial likelihood of the CPH model. Given the partial likelihood in Equation~(\ref{equ:likelihood}), we have defined $\mathcal{L}_\text{utility}$ as
\begin{equation}
	\mathcal{L}_\text{utility}= -\sum_{i:\delta_i=1}(\beta^\top x_i- \log\sum_{j:T_j\geq T_i}\exp(\beta^\top x_j))
\end{equation}
Next, we integrate Equation~(\ref{equ:FNDCG@k}) as the individual fairness regularizer $\mathcal{L}_\text{fairness}=\text{FNDCG@k}$ and define the unified objective function as
\begin{equation}
	\mathcal{L}_\text{unified}= \mathcal{L}_{\text{utility}}- \gamma\mathcal{L}_{\text{fairness}}
\end{equation}
\noindent where $\gamma$ is the tuning parameter controlling the trade-off between utility and fairness.
Combining $\mathcal{L}_{\text{utility}}$ and $\mathcal{L}_{\text{fairness}}$ in the unified objective function enables the learned model to be both accuracy-driven and fairness-oriented. 

There are two hyper-parameters governing fairIndvCox: $\gamma$, the coefficient controlling the balance between utility and fairness, and $k$, the length of the ordered list in the computation of $\text{DCG}_{\text{Sim}(d_n)}$. Both parameters effect our algorithm as a trade-off between the predictive performance and individual fairness, as we show empirically in Section \ref{sec:experiment_gamma} and \ref{sec:experiment_k}.

\section{Experiments}
\label{sec:experiments}

We conduct experiments to evaluate the effectiveness of our fairIndvCox algorithm, conduct a comparison study on our Lipschitz-free bias quantification, and examine the trade-offs controlled by the algorithm's hyper-parameters.

\subsection{Datasets}
\label{sec:datasets}
We validate our model on four real-world censored datasets with socially sensitive concerns: i) The \emph{ROSSI} dataset pertains to persons convicted then released from Maryland state prisons, who were followed up for one year after release~\cite{fox2012rcmdrplugin}. ii) The landmark algorithmic unfairness \emph{COMPAS} dataset to predict recidivism from Broward County~\cite{angwin2016there}. iii) The \emph{KKBox} dataset from the WSDM-KKBox's Churn Prediction Challenge 2017~\cite{kvamme2019time}. iv) The \emph{Support} dataset of hospitalized patients from five tertiary care academic centers~\cite{knaus1995support}.
See Table~\ref{tab:dataset_info} for the statistics. Note that survival information is explicitly included in these datasets to account for censoring. 

\begin{table}
\begin{center}
    \caption{Summary of datasets used in experiments} 
        \vspace{-0.1cm}
        \setlength{\tabcolsep}{4pt}
	\small
	\begin{tabular}{cccccc}
		\toprule
		\diagbox[width=3.5cm]{Characteristics}{Dataset}  & ROSSI & COMPAS & KKBox & Support \\
		\midrule
		Sample \# 			& 432       & 10,325    & 2,814,735 &8,873  \\
		Censored \#		    & 318       & 7,558     & 975,834 &2,840   \\
		Censored Rate		& 0.736     & 0.732     & 0.347  &0.320 \\
		Feature \# 			& 9         & 14        & 18 & 14\\
		\bottomrule
	\end{tabular} 
	\label{tab:dataset_info}
        \vspace{-0.9cm}
\end{center}
\end{table}

\begin{table*}[!t]
        \setlength{\tabcolsep}{4pt}
	\centering
	\begin{tabular}{cccccc}
		\toprule
		Dataset & \diagbox[width=2.5cm]{Method}{Metrics} & FNDCG@10\% & C-index\% & Brier \ score\% & Time-dependent \ AUC\%\\
		\midrule
		\multirow{5}{*}{ROSSI} 
		& FDCPH  &  44.12    & 55.81 & 19.83 & 76.18\\
		& CPH     & 33.41   & 64.24 & 17.67&77.12\\
		& RSF      & 36.17   & 65.56 & 15.12&\textbf{79.32}\\
		& DeepSurv & 31.43    & \textbf{66.67} & \textbf{14.71}&78.18 \\
		& \multirow{2}{*}{fairIndvCox}    & \textbf{53.29}     & 63.78 & 15.12 &78.25\\
		&              & \textbf{(20.78\%)}   & (-4.34\%)& (-2.79\%)& (-1.35\%)\\
		\midrule
		\multirow{5}{*}{COMPAS}  
		& FDCPH     & 72.27  & 63.54 &   24.12& 65.16 \\
		& CPH 		& 73.51  & 69.24   & 20.35& 65.15\\
		& RSF 		 & 74.64 & 72.61   & 15.62& 71.76\\
		& DeepSurv   & 74.18	& \textbf{75.12}   & \textbf{13.42} & 71.83\\
		& \multirow{2}{*}{fairIndvCox}	& \textbf{83.14}   & 68.73& 13.97 & \textbf{71.87}   \\
		&              & \textbf{(11.39\%)}   & (-8.50\%) & (-4.10\%) & (0.05\%) \\
		\midrule
        \multirow{5}{*}{KKBox} 
  & FDCPH   &  58.64 & 70.44  & 21.23 & 69.73 \\
  & CPH      & 47.32 & 80.02 & 18.17 & 72.95\\
  & RSF      & 42.41 & 82.32 & \textbf{14.24} & 78.18\\
  & DeepSurv  & 43.45 & 83.01 & 14.33 & 80.71 \\
  & \multirow{2}{*}{fairIndvCox}    & \textbf{67.44} & \textbf{83.27} & 14.45 & \textbf{80.95}\\
  &              & \textbf{(15.01\%)}   & (0.31\%) & (-1.47\%)& (0.29\%)  \\
  \midrule
        \multirow{5}{*}{Support}  
  & FDCPH     & 62.31  & 67.88 &   30.54& 76.34 \\
  & CPH   & 55.78  & 74.11   & 21.21& 80.02\\
  & RSF    & 65.15 & 75.18   & 16.64& \textbf{81.01}\\
  & DeepSurv   & 54.33 & \textbf{75.65}   & \textbf{16.11} & 80.68\\
  & \multirow{2}{*}{fairIndvCox} & \textbf{72.17}   & 74.31& 17.13 & 79.51   \\
  &              & \textbf{(10.78\%)}   & (-1.16\%) & (-6.33\%) & (-1.85\%) \\
        \midrule
	\end{tabular}
	\caption{Evaluation results of different models with the best results marked in bold. The numbers in parentheses represent the relative performance improvement of fairIndvCox compared to the best baseline.} 
	\label{tab:effectiveness}
        \vspace{-0.3cm}
\end{table*}

\subsection{Experiment Setup}
\paragraph{Baselines}
We compare fairIndvCox against four baselines: 
i) the recently proposed fair survival model FDCPH~\cite{keya2021equitable}, which, to the best of our knowledge, is the only work for fair survival analysis problem across individuals, 
ii) 
the classic survival analysis model CPH~\cite{cox1972regression}, 
iii) the state-of-the-art random forests modeling censored data RSF~\cite{ishwaran2008random}, and iv) the deep neural network on survival analysis DeepSurv~\cite{katzman2018deepsurv}. 
Other competing fairness methods are not considered as none of them is capable of addressing fairness in the presence of censoring. Neither are group-based fair survival models as they necessitate the specification of sensitive attribute to enforce fairness, which is absent in individual fairness learning. 

\paragraph{Predictive Performance Measures}
In addition to the proposed individual fairness measure, we also follow~\cite{zhang2022longitudinal} to report 
survival analysis metrics including 
C-index, Brier score, and time-dependent AUC as measures of predictive performance under censorship.
The C-index \cite{harrell1982evaluating} evaluates the probability that the predicted event-times of two individuals have the same relative order as their true event-times.
The Brier score \cite{brier1951verification} measures the mean squared difference between the predicted probability of outcome assignments and the true outcomes (the lower the Brier score, the better the prediction).
The time-dependent AUC \cite{chambless2006estimation} quantifies the probability that a randomly selected 
pair of individuals having experienced the event and not having experienced the event 
at time $t$
are correctly ordered.  

Without loss of generality, we employ the Euclidean distance with feature scaling to obtain $\text{Sim}_{D'}$. All methods are trained in the same way with 5-fold cross validation for fair comparison.
The Adam optimizer is used to optimize the model via backpropagation and automatic differentiation in PyTorch, with a learning rate of 0.01. The training is done in mini-batches of size 128 for 50 epochs. The overall objective function for quantitative performance comparison has top $k$ set as 10 and $\gamma$ set as 1. The base model Cox's hyperparameter settings are followed for the hidden unit number, and a grid search is conducted for fairness-specific tuning parameters. The search space for $k$ is 4-50 and for $\gamma$ it is $e^{-4}$ and $e^4$.

\subsection{Experiment Results}
Table~\ref{tab:effectiveness} shows the results of our experiment. Our new fairIndvCox clearly dominates all other baselines in minimizing discrimination (measured by FNDCG@10) while maintaining a competitive predictive performance (measured by C-index, Brier score, and time-dependent AUC), which verifies the necessity of its debiasing design across individuals while accounting for censorship. 
In contrast, the compared methods suffer from the lack of considering censored data as well as the non-trivial handling of the Lipschitz constant. 
The improved overall predictive performance of fairIndvCox also shows the merit of such an anti-discrimination design for prediction accuracy, presumably due to fairness regularization reducing overfitting.   

\vspace{-2cm}

\subsection{Comparison Study on the Lipschitz-free Bias Quantification}

We further perform a comparison study to demonstrate our method's advantage brought by being free from the Lipschitz condition in Equation~(\ref{equ:lipschitz}) which requires the specification of the Lipschitz constant during fairness quantification. We replace the $\mathcal{L}_\text{fairness}$ in fairIndvCox with Equation~(\ref{equ:lipschitz}) as suggested in~\cite{dwork2012fairness}, and denote the method as \textit{fairIndvCox-}. Results in Table~\ref{tab:confusionmatrix} show that fairIndvCox outperforms fairIndvCox- in minimizing discrimination for all datasets by large margins and also in terms of the predictive performance, except for a small decrease in the ROSSI dataset. This verifies that relaxing the Lipschitz constant specification in the conventional individual fairness definition can lead to improved performance.

\begin{table}
\begin{center}
        \setlength{\tabcolsep}{2.5pt}
	\small
	\centering
 	\caption{Results of comparison study on the Lipschitz-free Bias Quantification.} 
	\begin{tabular}{ccccccc}
		\toprule
		\multirow{2}{*}{Dataset}& \multicolumn{2}{c}{FNDCG@10\%} & \multicolumn{2}{c}{C-index\%}\\ 
		&   fairIndvCox-  & fairIndvCox  &fairIndvCox- & fairIndvCox   \\
		\midrule
		ROSSI   	 & 45.29    & 53.29    & 64.42  & 63.78 \\
		COMPAS   & 77.39   &  83.14   &  60.14  & 68.73 \\
		KKBox      &  54.02    & 67.44  &  82.71  & 83.27 \\
        Support    &  58.49    & 72.17  &  69.28  & 74.31 \\
		\bottomrule
	\end{tabular}
    \vspace{-0.3cm}
	\label{tab:confusionmatrix}
\end{center}
\end{table}



\begin{figure}[!t]
	\centering
	\includegraphics[width=0.5\textwidth]{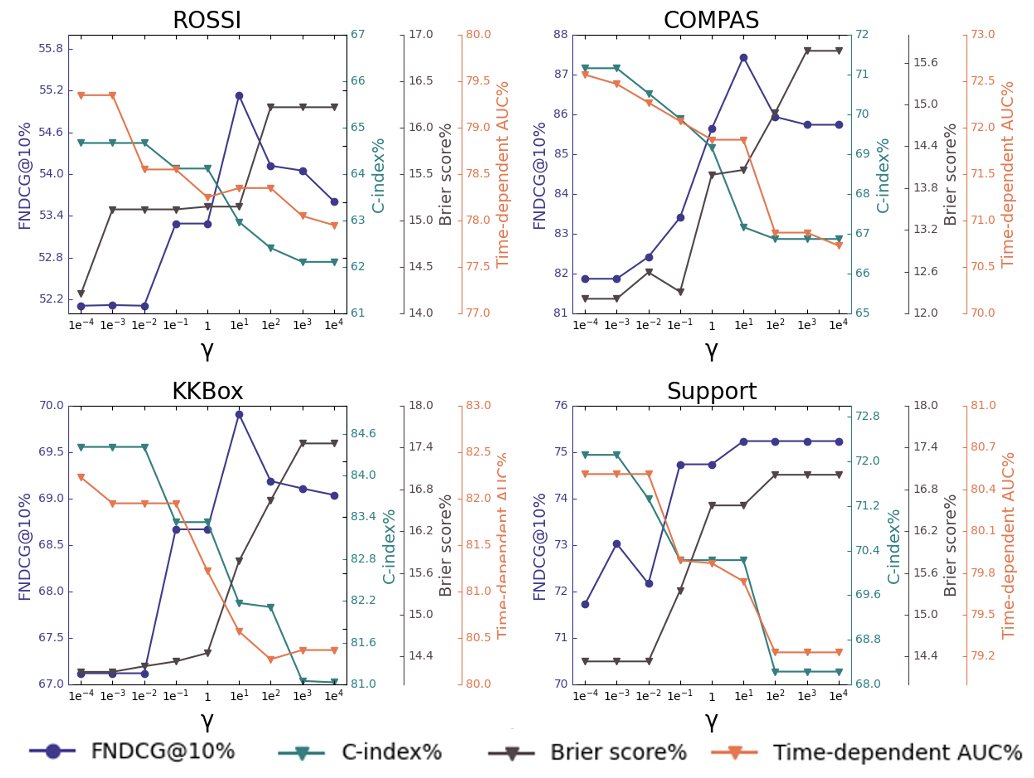}
	\caption{Study on individual fairness and accuracy trade-off on $\gamma$: The fairIndvCox models subject to different $\gamma$ variations (between $e^{-4}$ and $e^4$ ) on ROSSI, COMPAS, KKBox, and Support exhibit effects on individual fairness and model accuracy.}
    
 \label{fig:fairCalibrationGamma}
\end{figure}

\subsection{Effect of Different $\gamma$ Values on Individual Fairness and Predictive Performance}
\label{sec:experiment_gamma}
\sloppy
To investigate the effect of $\gamma$ on the performance of fairIndvCox, we vary $\gamma$ within the set $\{e^{-4}, e^{-3}, \cdots, e^4\}$ where $e$ is the natural constant, keeping all other hyper-parameters unchanged. We compare fairIndvCox's performance in terms of predictive power and individual fairness under the different settings.

According to the results shown in Figure~\ref{fig:fairCalibrationGamma}, there are three cases of $\gamma$ values. (1) For small $\gamma$ (\textit{i.e.}, less than $e^{-2}$ for ROSSI, $e^{-3}$ for COMPAS, $e^{-2}$ for KKBox, and $e^{-3}$ for Support), the individual fairness constraint has a small effect on fairIndvCox's predictive performance metrics (C-index, Brier score, and time-dependent AUC) and FNDCG@10 for the four tasks. (2) As $\gamma$ increases progressively (\textit{e.g.}, from $e^{-2}$ to $e^{1}$ for ROSSI, $e^{-3}$ to $e^{1}$ for COMPAS, $e^{-2}$ to $e^{1}$ for KKBox, and $e^{-3}$ to $e^{1}$ for Support), fairness increases significantly but at the cost of some predictive performance degradation (decreased C-index, decreased time-dependent AUC, and increased Brier score). This would imply that fairIndvCox achieves the appropriate balance between fostering individual fairness and preserving model performance. (3) When $\gamma$ is relatively large (\textit{e.g.}, larger than $e^{1}$ for all the datasets), the promotion of individual fairness will continue to have an effect on the predictive performance, with the exception of the Support dataset where both FNDCG@10 and the predictive performance metrics stay mostly fixed when $\gamma$ is greater than $e^2$. Note that FNDCG@10 mostly also decreases as $\gamma$ increases since it adds more weight to $\mathcal{L}_{\text{fairness}}$. But this does not mean we can obtain the optimal node. 
Therefore, the performance of individual fairness promotion within a fixed epoch is close to its limit, and it is difficult to achieve better performance.

\begin{figure}[!ht]
	\centering
	\includegraphics[width=0.5\textwidth]{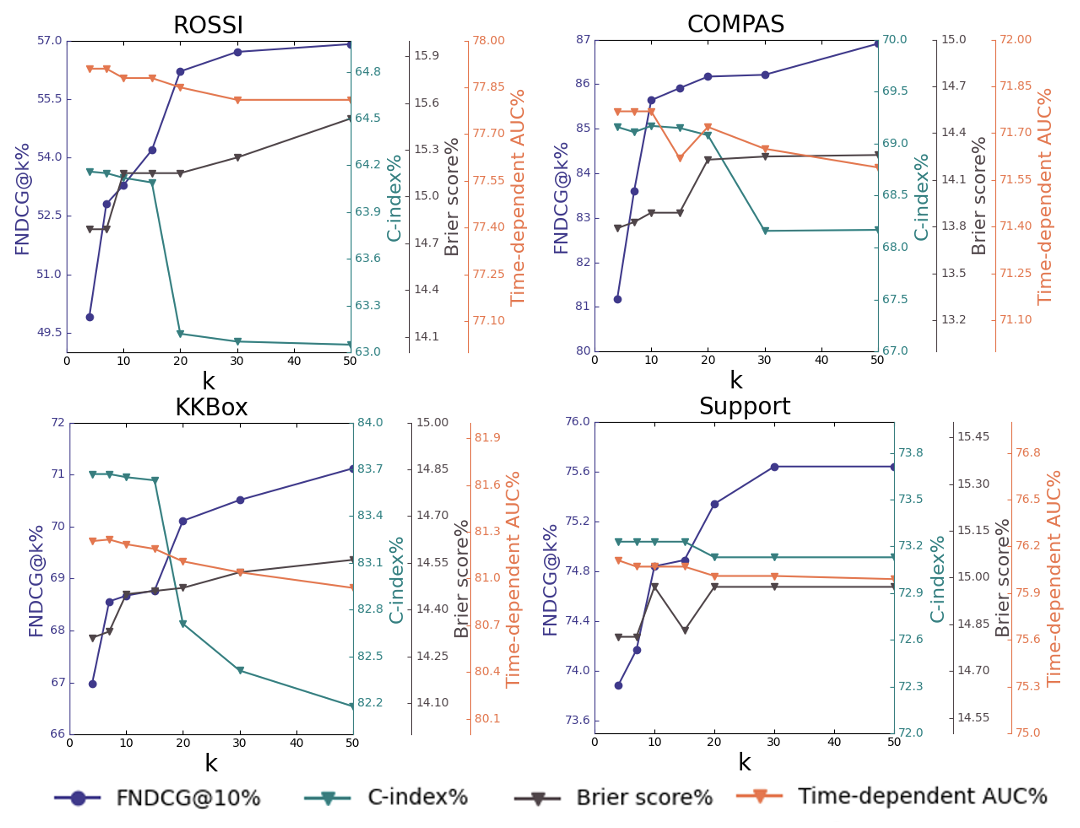}
	\caption{Study the choice of k-value: The fairIndvCox models subject to different k variations (between 4 and 50) on ROSSI, COMPAS, KKBox, and Support exhibit effects on individual fairness and model accuracy.}
	\label{fig:fairCalibrationK}
\end{figure}

\subsection{Effect of Different Number of Neighbors $k$ Values on Individual Fairness and Predictive Performance}
\label{sec:experiment_k}

Similar to the previous section, we conducted experiments with a variety of values for $k$ in $\{4, 7, 10, 15, 20, 30, 50\}$, keeping all other training factors the same. We compare fairIndvCox's predictive performance and fairness under different settings.

We observe that (Figure~\ref{fig:fairCalibrationK}): (1) As $k$ increases, the fairIndvCox achieves better performance on \textit{FNDCG@k}, demonstrating better optimization for individual fairness. (2) When $k$ is a modest value (e.g., smaller than 15 for ROSSI, 20 for COMPAS, 15 for KKBox, and 10 for Support), the predictive performance (as measured by C-index and time-dependent AUC) is hardly affected or even increases, though the Brier score performs slightly worse (increased). The fairIndvCox mostly strikes the right balance between maintaining model utility and fostering individual fairness with proper choices of $k$ in here. (3) When $k$ is significant (e.g., greater than 15 for ROSSI, 20 for COMPAS, 15 for KKBox, and 10 for Support), the predictive performance significantly declines (decreased C-index and time-dependent AUC, and increased Brier score), with the exception on the Support dataset where the Brier score fluctuates when k is between 10 and 20, and all three metrics stay relatively flat when k is greater than 20. In general, more points will be referenced at a time as $k$ increases, resulting in more interference values. This leads to a decrease in the weight of the correct label and a blurred classification, causing degradation in predictive performance.
\vspace{-0.2cm}

\section{Conclusion}
\label{sec:conclusion}

A striking gap exists between the prevailing real-world applications with censorship and the assumption of full class label availability in existing AI fairness methods. We make an initial investigation of individual fairness in learning with censorship. Moreover, this work defines individual fairness from a ranking perspective, relaxing from the Lipschitz condition in conventional individual fairness studies. The proposed notion and algorithm are expected to be versatile in quantifying and mitigating bias in various socially sensitive applications. We provide an empirical evaluation of four real-world datasets to validate the effectiveness of our method. The experimental results show that with suitable $\gamma$ and $k$ values, our method can substantially improve individual fairness with an acceptable loss of predictive performance as the model outperforms the current state-of-the-art individual fairness promotion methods. Finally, this work defines a new task that opens up possibilities for future work to achieve more applicable and comprehensive AI fairness. 

\section*{Acknowledgement}

This work was supported in part by National Science Foundation (NSF) under Grant No. 2245895 and an NVIDIA GPU Grant. 

\appendix





\bibliography{ijcai23, typeinst}

\end{document}